\documentclass[smallextended]{svjour3}       

\usepackage{amsbsy}
\usepackage{amsmath}
\usepackage{amssymb}
 \usepackage{float}
\usepackage{booktabs}

\usepackage{fullpage}
\usepackage{bm}
\usepackage{mathrsfs}
\usepackage{enumerate}
\usepackage{multirow}
\usepackage{multicol}
\usepackage{array}
\usepackage{color}
\usepackage{epsfig}
\usepackage{graphicx}	
\usepackage{subfigure}
\usepackage{booktabs}
\usepackage{times}

\usepackage{algorithm}  
\usepackage{algorithmicx}  
\usepackage{algpseudocode}

\usepackage{booktabs}
\usepackage{threeparttable}
\usepackage[titletoc]{appendix}
\usepackage[T1]{fontenc}
\usepackage[utf8]{inputenc}

\usepackage[colorlinks]{hyperref}

\usepackage{authblk}
\usepackage{natbib}

\makeatletter
\def\singlespace{\def\baselinestretch{1}\@normalsize}

\renewcommand{\hat}{\widehat}

\begin{document}

\title{ASBART:Accelerated Soft Bayes Additive Regression Trees}

\author{Ran Hao         \and
	Bai Yang\thanks{Corresponding author: email@mail.com}
}

\institute{Ran Hao \at
	Department of applied statistics,Shanghai University of Finance and Economics,China \\
	\and
	Bai Yang \at
	Department of applied statistics,Shanghai University of Finance and Economics,China\\}           

\date{}
\maketitle

\begin{abstract}

Bayes additive regression trees(BART) is a nonparametric regression model which has gained wide -spread popularity in recent years due to its flexibility and high accuracy of estimation.
Soft BART,one variation of BART,improves both practically and theoretically on existing Bayesian sum-of-trees models.
One bottleneck for Soft BART is its slow speed in the long MCMC loop.
Compared to BART,it use more than about 20 times to complete the calculation with the default setting.
We proposed a variant of BART named accelerate Soft BART(ASBART).
Simulation studies show that the new method is about 10 times faster than the Soft BART with  comparable accuracy.
Our code is open-source and available at \url{https://github.com/richael008/XSBART}.
\end{abstract}



\baselineskip=20pt
\section{Introduction}
\label{sec:1}
\indent
\vspace{-5mm}

Suppose we have a response Y and $p$-dimensional predictor X of n samples.
 Consider the regression model
\begin{eqnarray}
	Y_{i}=f_{0}(X_{i})+\varepsilon_{i},i=1, \cdots, n
\end{eqnarray}
where  Gaussian noise $\varepsilon_{i} \sim N\left(0, \sigma^{2}\right) $
and $f_{0}$ is an unknown function of interest.
The task is to set up a nonparametric model that can capture relationships $f_{0}$ between X and Y.

Bayesian Additive Regression Trees(BART)\citep{chipman2010bart} is a nonparametric regression model that is often more accurate than other tree-based methods as random forest\citep{breiman2001random},Xgboost\citep{chen2016xgboost}.
It loose some stringent parametric assumptions compared to other parametric model and combines the flexibility of a machine learning algorithm with the formality of likelihood-based inference to create a powerful inferential tool.
Another advantage for BART model is its robust performance with respect to various hyperparameter settings so we don't have to waste much time on tuning hyperparameter to achieve a perfect fitting.

A problem shared with other tree models is that the resulting estimates of model are step functions,which can introduce error into the model.
The BART model archive some degree of smoothing by averaging over the posterior distribution.
If the underlying $f_{0}(X)$ is smooth,we can take advantage of this additional smoothness to get a more accurate model.
\citet{linero2018bayesianb} introduce smoothness to BART model by changing the decisions made at each node as random rather than deterministic.
For example,sample x goes right at branch b of tree $\mathcal{T}$  with probability $\psi(x ; \mathcal{T}, b)=\psi(x ; c_{b}, \tau_{b})=\psi\left(\frac{x_{j}-c_{b}}{\tau_{b}}\right)$  where
$\tau_{b}> 0$ is the bandwidth parameter and $c_{b}$ is the splitting value associated with branch b,$x_{j}$ is the splitting variable.
We usually set
\begin{eqnarray}
	\psi(x ; c_{b},\tau_{b})=\left(1+e^{-(x-c_{b}) / \tau_{b}}\right)^{-1}
\end{eqnarray}
so that smaller values of x will have higher probability of going left and vice versa.
Note that when $\tau_{b}\rightarrow 0$ ,the random decision is equal to the deterministic decision of BART.
\citet{linero2018bayesianb} refer to trees constructed using the above random decision rule as soft trees and call this BART variant as Soft BART(SBART).
They also showed the substantial theoretical and practical benefits for SBART.
The primary drawback is that it needs to compute every node's $\psi(x ; \mathcal{T}, b)$ for every sample x,rather than just a single leaf node in BART model.
They mentioned that the SBART is the slowest package among the competitors.
Actually the slowness impede the application of SBART.

\citet{he2019xbart} develops a modified version of BART that is amenable to fast posterior estimation and matches the previous BART and is many times faster and less memory intensive. 

The main contributions of this paper can be summarised as follows:

 \begin{itemize}
	\item  Combine the SBART and XBART together to overcome the shortcoming of SBART.
	\item  The experimental results demonstrate that the proposed method can accelerate the SBART process with a comparable accuracy. 
\end{itemize}

The paper proceeds as follows: Section \ref{sec:2} reviews BART, including SBART and XBART. 
In Section \ref{sec:3} we introduce ASBART in detail.  
Some experiments are conducted in section \ref{sec:4} to examine and compare ASBART with BART. 
Finally, section \ref{sec:5} offers the conclusions of the paper as well as some future works.

\section{Bayesian additive regression trees (BART)}
\label{sec:2}
\subsection{Model}
\label{sec:2.1}
This section motivates and describes the BART framework. We begin our discussion from a basic BART with independent continuous outcomes, because this
is the most natural way to explain BART.

For data with n samples ,the $i^{th}$ sample is consist of a p-dimensional vector of predictors $X_{i}$ and a response $Y_{i} (1\leq i \leq n)$, the BART model posits
\begin{eqnarray}
	Y_{i}=f(X_{i})+\varepsilon_{i},  \varepsilon_{i} \sim N\left(0, \sigma^{2}\right) ,i=1, \cdots, n.
\end{eqnarray}
To estimate $f(X)$, a sum of regression trees is specified as

\begin{eqnarray}
	f(X_{i})=\sum_{j=1}^{m} g\left(X_{i} ; T_{j}, M_{j}\right),
\end{eqnarray}
where $T_{j}$ is the $j^{th}$ binary tree structure and  $M_{j}=\left\{\mu_{1 j}, \ldots, \mu_{b_{j}}\right\}$is the terminal node parameters associated with $T_{j}$. $T_{j}$ contains information of which bivariate to split on, the cutoff value, as well as the internal node's location. The hyperparameter number of trees m is usually set as 200.

\subsection{Prior }
\label{sec:2.2}
BART is based on Bayes theory. The prior distribution for BART is denoted as $P\left(T_{1}, M_{1}, \ldots, T_{m}, M_{m}, \sigma\right)$. Here we assume that $\left\{\left(T_{1}, M_{1}\right), \ldots,\left(T_{m}, M_{m}\right)\right\}$ are independent with $\sigma$, and $\left(T_{1}, M_{1}\right), \ldots,\left(T_{m}, M_{m}\right)$ are also independent with each other, so we have
\begin{eqnarray}
	\label{equ:s1}
	\begin{aligned}
		P\left(T_{1}, M_{1}, \ldots, T_{m}, M_{m}, \sigma\right) &=P\left(T_{1}, M_{1}, \ldots, T_{m}, M_{m}\right) P(\sigma) \\
		& =\left[\prod_{j}^{m} P\left(T_{j}, M_{j}\right)\right] P(\sigma) \\
		&=\left[\prod_{j}^{m} P\left(M_{j} \mid T_{j}\right) P\left(T_{j}\right)\right] P(\sigma) \\
		&=\left[\prod_{j}^{m}\left\{\prod_{k}^{b_{j}} P\left(\mu_{k j} \mid T_{j}\right)\right\} P\left(T_{j}\right)\right] P(\sigma).
	\end{aligned}
\end{eqnarray}
From $(\ref{equ:s1})$, we need to specify the priors of $P\left(\mu_{k j} \mid T_{j}\right)$, $P(\sigma)$, and $ P\left(T_{j}\right)$ respectively. For the convenience of computation, we use the conjugate normal distribution $N\left(\mu_{\mu}, \sigma_{\mu}^{2}\right)$ as the prior for $\mu_{i j} \mid T_{j}$. The initial prior parameter $(\mu_{\mu}$, and $\sigma_{\mu})$ can be set  through  roughly computation. 
We also use a conjugate prior, here the inverse chi-square distribution for $\sigma$, $\sigma^{2} \sim v \lambda / \chi_{v}^{2}$, where the two hype-parameters $\lambda$, $v$ can be roughly derived by calculation.
The prior for $T_{j}$ is specified and made up of three aspects:
\begin{itemize}
	\item [1)]  The probability for a node at depth $d$ to split: given by $\frac{\alpha}{(1+d)^{\beta}}$. We can confine the depth of each tree by controlling the splitting probability so that we can avoid overfitting. Usually $\alpha$ is set to 0.95 and $\beta$ is set to 2.
	\item [2)]  The probability on splitting variable assignments at each interior node: default as uniform distribution. Dirichlet distribution are introduced for high dimension variable selection scenario \citep{linero2018bayesiana, linero2018bayesianb}.
	\item [3)]  The probability for cutoff value assignment: default as uniform distribution .
\end{itemize}

\subsection{Posterior Distribution}
\label{sec:2.3}

With the settings of priors $(\ref{equ:s1})$, the posterior distribution can be obtained by
\begin{eqnarray}
	\label{equ:s2}
	\begin{aligned}
		P\left[\left(T_{1}, M_{1}\right), \ldots,\left(T_{m}, M_{m}\right), \sigma \mid Y\right] \propto & P\left(Y \mid\left(T_{1}, M_{1}\right), \ldots,\left(T_{m}, M_{m}\right), \sigma\right) \\
		& \times P\left(\left(T_{1}, M_{1}\right), \ldots,\left(T_{m}, M_{m}\right), \sigma\right) ,
	\end{aligned}
\end{eqnarray}
where $(\ref{equ:s2})$ can be obtained by  Gibbs sampling. First m successive
\begin{eqnarray}
	\label{equ:s3}
	\begin{aligned}
		P\left[\left(T_{j}, M_{j}\right) \mid T_{(j)}, M_{(j)}, Y, \sigma\right]
	\end{aligned}
\end{eqnarray}
can be drawn where $T_{(j)}$ and $M_{(j)}$ consist of
all the trees information except the $j^{th}$ tree. Then explicit inverse gamma distribution of $P\left[    \sigma \mid \left(T_{1}, M_{1}\right), \ldots,  Y\right]$ can be obtained.

How to draw from $(\ref{equ:s3})$ ? $T_{j}$, $M_{j}$ depends on  $ T_{(j)}$, $M_{(j)}$ and  $Y$ through
$R_{j}=Y-\sum_{w \neq j} g\left(X, T_{w}, M_{w}\right)$
, it is equivalent to draw posterior from a single tree of
\begin{eqnarray}
	\label{equ:s4}
	P\left[\left(T_{j}, M_{j}\right) \mid R_{j}, \sigma\right].
\end{eqnarray}
$(\ref{equ:s4})$  is split in two steps. First draw posterior from $P\left(T_{j} \mid R_{j}, \sigma\right)$, then draw posterior from $P\left( M_{j} \mid T_{j},  R_{j}, \sigma\right)$.
In the first step, we have
\begin{eqnarray} \label{equ:s5}
	P\left(T_{j} \mid R_{j}, \sigma\right) \propto P\left(T_{j}\right) \int P\left(R_{j} \mid M_{j}, T_{j}, \sigma\right) P\left(M_{j} \mid T_{j}, \sigma\right) d M_{j},
\end{eqnarray}
we call $P\left(R_{j} \mid T_{j}, \sigma\right)= \int P\left(R_{j} \mid M_{j}, T_{j}, \sigma\right) P\left(M_{j} \mid T_{j}, \sigma\right) d M_{j}$ as marginal likelihood. Because  conjugate Normal prior is employed on $ M_{j}$, we can get an explicit expression of the marginal likelihood

\begin{equation}
	\label{eq14}
	\begin{split}
		\sum_{b=1}^{B}\{&-\frac{n_{b}}{2} \log (2 \pi)-n_{b} \log (\sigma)+\frac{1}{2} \log \left(\frac{\sigma^{2}}{\sigma^{2}+\tau n_{b}}\right)\left.-\frac{1}{2} \frac{y_{b}^{t} y_{b}}{\sigma^{2}}+\frac{1}{2} \frac{\tau}{\sigma^{2}\left(\sigma^{2}+\tau n_{b}\right)} s_{b}^{2}\right\} \\
		=&-n \log (2 \pi)-n \log (\sigma)-\frac{1}{2} \frac{y^{t} y}{\sigma^{2}} + \frac{1}{2} \sum_{b=1}^{B}\left\{\log \left(\frac{\sigma^{2}}{\sigma^{2}+\tau n_{b}}\right)+\frac{\tau}{\sigma^{2}\left(\sigma^{2}+\tau n_{b}\right)} s_{b}^{2}\right\}
	\end{split}
\end{equation}
where $\sigma^{2}$is the model variance,$\tau$ is the mean variance of the tree model,$B$ is the number of the leaves node, $n_{b}$is the valid samples in the $b^{th}$ leaf node, $S_{b}$is the sum of response variable in that leaf node,$y$ is the vector of  response variable.

We process $(\ref{equ:s5})$ by generating a candidate tree $T_{j}^{*}$ from the previous tree structure with  MH algorithm.
we accept the new tree structure with probability

\begin{eqnarray} \label{equ:s61}
	\alpha\left(T_{j}, T_{j}^{*}\right)=\min \left\{1, \frac{q\left(T_{j}^{*}, T_{j}\right)}{q\left(T_{j}, T_{j}^{*}\right)} \frac{P\left(R_{j} \mid X, T_{j}^{*}\right)}{P\left(R_{j} \mid X, T_{j}\right)} \frac{P\left(T_{j}^{*}\right)}{P\left(T_{j}\right)}\right\},
\end{eqnarray}
Where $q\left(T_{j}, T_{j}^{*}\right)$ is  the probability for the previous tree $T_{j}$ moves to the new tree $T_{j}^{*}$.

The candidate tree is growing from previous tree structure by randomly selecting one of the four type of modifications:Grow,Prune,Swap and Change.

Once we have finished sample from  $P\left(T_{j} \mid R_{j}, \sigma\right)$, we can sample the $j^{th}$ leaf parameter $\mu_{k j}$ of  the $k^{th}$ tree  from  posterior Normal distribution.
We repeat this process for many iterations and drop numbers of first unstable iterations and finally keep the stable iterations as the non-parameter estimation.

\subsection{SBART and XBART}
\label{sec:2.4}

\cite{linero2018bayesianb} introduce the Soft Bayesian additive regression trees(SBART) model, which uses soft decision trees which effectively replace the decision boundaries of BART with smooth sigmoid functions.

With the definition of the  logistic gating function $\psi(x)$,the probability of going to leaf $\ell$  is
\begin{eqnarray}
	\phi(x ; \mathcal{T}, \ell)=\prod_{b \in A(\ell)} \psi(x ; \mathcal{T}, b)^{1-R_{b}}(1-\psi(x ; \mathcal{T}, b))^{R_{b}}
\end{eqnarray}
where $A(\ell)$ is the set of ancestor nodes of leaf $\ell$ and $R_{b}=1$ if the path to $\ell$ goes right at $b$.Here we denote $\phi_{i}$ as the probability vector for the ith sample $x_{i}$ to go to each leaf of the tree.Besides the randomized decision rule,SBART use a sparsity-inducing Dirichlet  as prior distribution for splitting variables so it can adapted to high dimensional scenario for variable selection.

BART’s wider adoption has been slowed by its more severe computational demands relative to alternatives, owing to its reliance on a random walk Metropolis-Hastings Markovchain Monte Carlo (MCMC) approach.To overcome this problem, \cite{he2019xbart}  develop a modified version of BART,Accelerated Bayesian Additive Regression Trees(XBART) which is amenable to fast posterior estimation. They present a stochastic hill climbing algorithm that
matches the remarkable predictive accuracy of previous BART implementations,but is many times faster and less memory intensive.With the XBART algorithm.For each iteration their algorithm grow a tree from root which is quiet different from the original BART which try a small modification on the previous tree structure. 

Considering the good performance of the SBART and XBART,our method combine the two algorithm together.More detailed description later.

\section{Accelerated Soft Bayes Additive Regression Trees}
\label{sec:3}

\citet{he2019xbart} presents a method for accelerating Bayesian additive regression trees, known as accelerated Bayesian additive regression trees (XBART). By employing this method, it is possible to quickly achieve precision similar to Bayesian additive regression trees while expending only a fraction of the time, comparable to a few dozen times less than that of BART. The time required is equivalent to approaches like XGBoost, but with higher accuracy.

Diverging from the iterative process of Bayesian additive regression trees in the $k+1$ iteration, where a small step modification is attempted on the tree structure $T_{l}^{(k)}$ to generate a new structure $T_{l}^{(k)}$, XBART employs a mode of growth from the root node in each step. At each node, the marginal likelihood of all possible splits for (selected) variables is computed, and then a sampling process incorporating the ratios of these marginal likelihoods and the probabilities of stopping growth is used to decide whether to halt growth or proceed with the selected action. This iteration is executed at each node, resulting in a well-grown tree. During this growth process, only one action is taken, as opposed to Bayesian additive regression trees that involve multiple alteration scenarios. Precalculation of marginal likelihood results for various splits is facilitated by pre-sorting the data, allowing for rapid computation.

The reason for considering smoothed Bayesian additive regression trees lies in their superior ability to characterize continuous functions compared to BART. However, if obtaining marginal gains in estimation accuracy comes at a high cost, such as a several-fold increase in computational time, the cost-effectiveness might be diminished. The primary objective of this section is to integrate the principles of accelerated Bayesian additive regression trees to expedite the computational speed of SBART, thus enhancing their competitiveness.

Given the relatively sluggish computational speed of smoothed Bayesian additive regression trees, our plan involves integrating the methodology of XBART into the computational process of SBART. We intend to use the XBART approach to generate an alternative tree. However, considering that the trees derived from XBART correspond to estimates of traditional non-smooth trees, we will then seek to determine the optimal window width for the XBART-generated tree, thereby obtaining an estimate of a smooth tree.

Why can this process yield an estimate of a smooth tree? The rationale lies in our observation that the marginal likelihood of SBART is given by

\begin{equation}
	\label{eq4.4}
	P\left(R_{j} \mid T_{j}, \sigma, \sigma_{\mu}, W_{j}\right)=\frac{|2 \pi \Omega|^{1 / 2}}{\left(2 \pi \sigma^{2}\right)^{n / 2}\left|2 \pi \sigma_{\mu}^{2} I\right|^{1 / 2}} \exp \left(-\frac{\left\|R_{j}\right\|^{2}}{2 \sigma^{2}}+\frac{1}{2} \widehat{\mu}^{\top} \Omega^{-1} \widehat{\mu}\right)
\end{equation}	
where
\begin{equation}
	\Omega=\left(\frac{\sigma_{\mu}^{2}}{T} \mathrm{I}+\Lambda\right)^{-1}, \quad \Lambda=\sum_{i=1}^{n}  \phi_{i} \phi_{i}^{\top} / \sigma^{2}, \quad \widehat{\mu}=\Omega \sum_{i=1}^{n} R_{i} \phi_{i} / \sigma^{2}
\end{equation}

While the forms of (\ref{eq4.4}) and (\ref{eq14}) may differ significantly, in fact, (\ref{eq14}) can be seen as a special case of (\ref{eq4.4}) when the window width is 0. In other words, the two marginal likelihoods are exactly equal when the window width is 0. We assume that the objective function is a continuously differentiable function. When (\ref{eq14}) is relatively large, the corresponding (\ref{eq4.4}) after finding the optimal window width is also relatively large. Consequently, the posterior probability of this tree is high, increasing its chance of being selected. The specific algorithm can be referred to as Algorithm \ref{alg:5}.

\begin{algorithm}
	\caption{Accelerated Soft Bayes Additive Regression Trees}
	\label{alg:5}
	\begin{algorithmic}[1]
		\Procedure {XSBART}{$y$, $X$, $L$, $K$}
		\State initialize residential  $r_{0}^{0}\leftarrow y$
		\For {$k \leftarrow 1,\cdots, K$}
		\For {$l \leftarrow 1,\cdots, L$}
		\State compute residential $r_{l}^{k}$
		\State Grow candidate tree $T_{l}^{k}$ with GROWFROMROOT$(r_{l}^{k},X)$ from XBART
		\State Find optimal band width for $T_{l}^{k}$
		\State MH sampling to decided to choose $T_{l}^{k}$ or keep $T_{l}^{k-1}$
		\EndFor		
		\EndFor
		\EndProcedure
	\end{algorithmic}
\end{algorithm}

Several distinctive parts of this algorithm, differing from other algorithms, will be explained below.

\textbf{Optimal Window Width Search}

In this context, we utilize a grid-point search to determine the optimal window width. For instance, we perform the search over grid points $(0\%, 1\%, 2\%, \ldots, 20\%)$, where the $p\%$ grid point indicates that the range of the selected split point includes $2p\%$ of the population's samples. This approach differs slightly from SBART, where the independent variables are preprocessed into percentiles, allowing the direct use of $p\%$ as the window width criterion. Another distinction lies in SBART's assignment of an exponential prior distribution to the window width, followed by MH sampling to decide whether to accept a new bandwidth.

The default approach is to use a uniform window width for each tree. An alternative option is to search for the window width at each node, which entails a significant computational burden. Each window width search requires a global data computation, and from the results, it's evident that this method doesn't effectively enhance data accuracy. Instead, it consumes substantial computational resources. Therefore, in subsequent steps, we continue to adopt the approach of using a uniform window width for individual trees.

\textbf{MH Tree Sampling}

When we employ XBART to obtain tree estimates, we don't need as many iterations for trees as BART. In such cases, as each tree is generated by growing from a single root node, there's a possibility that previously obtained good estimates might be discarded. This operation, in fact, doesn't facilitate accuracy improvement. We found that utilizing MH sampling can help make better use of tree information.

Using XBART's growth-from-root method, we generate an alternative tree structure $T_j^*$. Subsequently, we compare the alternative tree structure $T_j^*$ with the tree structure from the previous iteration, $T_j$. Then, using MH sampling, we determine whether to accept the new tree structure $T_j^*$ or retain the previous one $T_j$, based on a certain probability

\begin{eqnarray} \label{equ:s6}
	\alpha\left(T_{j}, T_{j}^{*}\right)=\min \left\{1, \frac{q\left(T_{j}^{*}, T_{j}\right)}{q\left(T_{j}, T_{j}^{*}\right)} \frac{P\left(R_{j} \mid X, T_{j}^{*}\right)}{P\left(R_{j} \mid X, T_{j}\right)} \frac{P\left(T_{j}^{*}\right)}{P\left(T_{j}\right)}\right\}.
\end{eqnarray}

We choose to accept the alternative tree structure $T_j^{*}$. In this context, $q(T_j^{*}, T_j)$ and $q(T_j, T_j^{*})$ represent the transition probabilities between the two tree structures. $P(R_j | X, T_j^{*})$ and $P(R_j | X, T_j)$ are the marginal likelihoods for the two tree structures, while $P(T_j^{*})$ and $P(T_j)$ denote the priors for the respective tree structures. Unlike BART, both trees $T_j^{*}$ and $T_j$ are grown from root nodes, so $\frac{q(T_j^{*}, T_j)}{q(T_j, T_j^{*})} = 1$.
Furthermore, the prior probabilities of the two tree structures, due to the growth process traversing all possible splits, are independent of the split points and split variables and are only related to the splitting probabilities. Through this step of MH sampling, we strive to retain tree structures with higher posterior probabilities, while also preserving the potential for exploring tree structures with lower posterior probabilities.

\textbf{Ordinal Variables}

Independent variables can be categorized into categorical variables and ordinal variables. For categorical variables, setting a window width is not meaningful. Although SBART treats categorical variables and ordinal variables similarly, in our approach, if a tree contains only ordinal variables, there's no need to search for a window width. This helps reduce computational complexity.

XBART's treatment of categorical variables has some shortcomings. It treats categorical variables the same as ordinal variables, using split rules like "greater than or equal to a certain value." However, categorical variable values lack inherent meaning. If applied in this way, especially in the presence of multiple categorical variables, it could negatively impact estimation accuracy. SBART's approach involves transforming categorical variables into dummy variables, allowing a single split to filter out a specific category. This treatment is more reasonable than XBART's approach, making it akin to SBART's handling. Consequently, we adopt a similar approach of transforming categorical variables into dummy variables.

\textbf{Gate Function}

The original gate function in SBART was the Sigmoid function$$\psi((x-c)/\tau)=\left(1+e^{-(x-c)/\tau}\right)^{-1}.$$
Its advantage lies in approximating smooth functions by selecting different window widths. However, for data significantly distant from the split point, such as when $X > C + \tau$, the function's value gradually approaches 0 but never reaches it. These small values persist throughout node splitting, potentially having an exceedingly subtle impact on the final result, which might be negligible. To address this, a linear gate function is considered:

$$ \psi((x-c)/\tau)=\left\{
\begin{array}{rcl}
	0 & & {x-c<-\tau}\\
	(x-c)/2\tau +0.5 & & {-\tau \leq x-c \leq \tau }\\
	1 & & {x-c>\tau}
\end{array} \right . $$

\begin{figure}[htb]
	
	\begin{center}
		\includegraphics[width=0.7\textwidth]{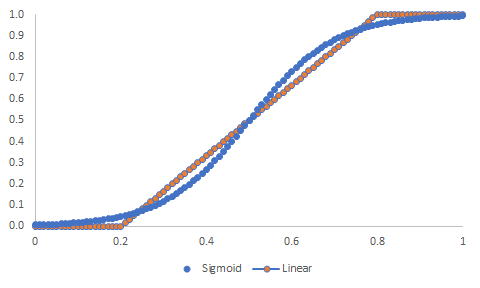}
	\end{center}
	\caption  {Comparation of the two Gating function}
	\label{FIG_Gate}
\end{figure}

Figure \ref{FIG_Gate} compares the function graphs of the Sigmoid gate function with $\tau = 0.1$ and the linear gate function with $\tau = 0.3$ when $c = 0.5$. It's evident that these two functions are quite close and capture the majority of the function's fluctuations.
The comparative analysis of computational efficiency and estimation accuracy under these two gate function settings for the accelerated smooth Bayesian model will determine whether an update to the gate function is warranted.

\section{Data Experiments}
\label{sec:4}
\subsection{Time-accuracy comparisons to other methods}
We illustrate the disparities in model prediction accuracy and time consumption between XSBART and various BART models through a small-scale data experiments. For this purpose, we have selected 20 independent variables denoted as $x=\left(x_{1}, x_{2}, \ldots, x_{20}\right)$, where
$x_{1}, x_{2}, \ldots, x_{20}$ i.i.d. $\sim \operatorname{Uniform}(-2,2)$. 
The Friedman five-dimensional test function is used to generate $y$:
$$y=f(x)+\varepsilon=10 \sin \left(\pi x_{1} x_{2}\right)+20\left(x_{3}-0.5\right)^{2}+10 x_{4}+5 x_{5}+\varepsilon$$
Two error settings are considered: high noise $\varepsilon \sim Var(f)$ or low noise $\varepsilon \sim 1$. 
The training and test dataset both consist of 2000 samples. 
Total 100 sets of data are generated.

Due to the intrinsic connection between the number of iterations, program execution time and estimation accuracy.
For SBART, we consider three settings:
\begin{itemize}
	\item The first 500 iterations are discarded as burn-in samples, and the subsequent 500 iterations are retained. This setting is denoted as S500.
	\item The first 1000 iterations are considered as burn-in samples, and the subsequent 1000 iterations are retained. This configuration is denoted as S1000.
	\item The first 2000 iterations are taken as burn-in samples, and the subsequent 2000 iterations are retained. This configuration is denoted as S2000.
\end{itemize}

XBART maintains the default settings of the original software package: a total of 40 iterations, with the first 15 iterations treated as burn-in samples. 
BART is configured such that the first 2000 iterations are considered as burn-in samples, and the subsequent 2000 iterations are retained.

First we commence by addressing the high noise setting $\varepsilon \sim Var(f)$.

\begin{figure}
	\caption{Comparation of different method in high noise scenario}
	\label{FIG4-1}
	\begin{center}
		\includegraphics[width=0.8\textwidth]{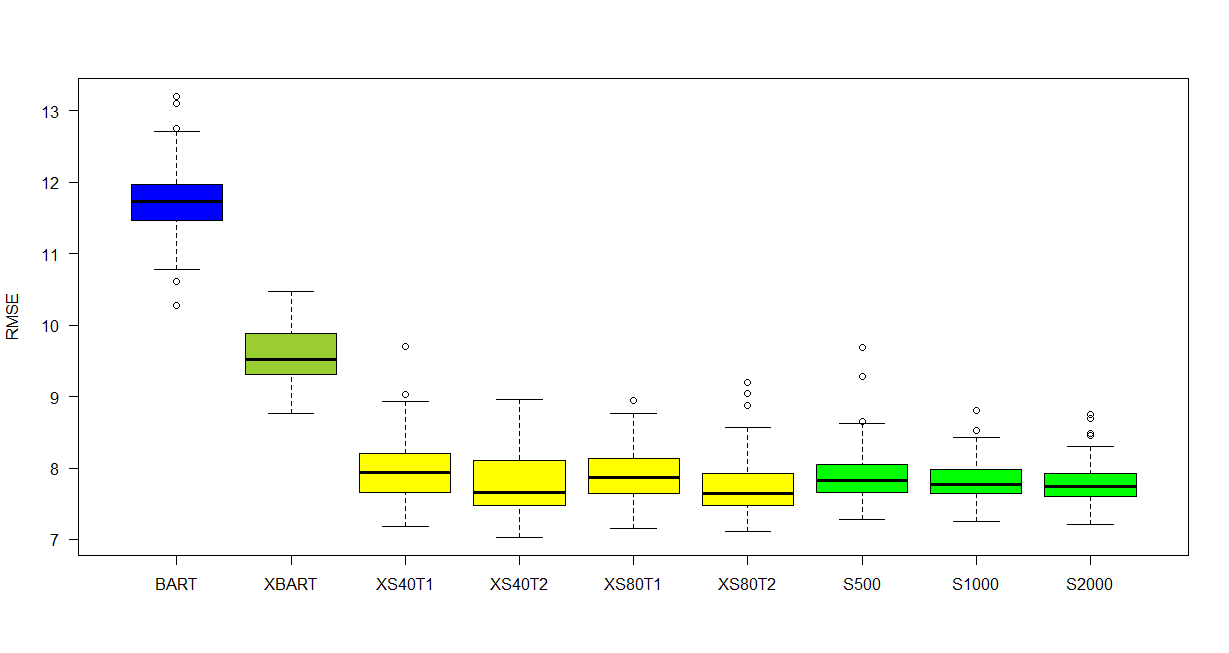}
	\end{center}
	\small{\it{XS40 stands for XSBART with total of 40 iterations. T1 represents the utilization of a linear gate function for estimation, while T2 signifies the application of a sigmoid gate function. S500 denotes SBART which  discard  the first 500 iterations and keep the subsequent 500 iterations. Similar naming apply for the remaining configurations.}}
\end{figure}

The statistical measure we employ here to assess model prediction capability is $$\mathrm{RMSE}=\sqrt{\frac{1}{n} \sum_{i=1}^{n}\left(\hat{f}\left(x_{i}\right)-f\left(x_{i}\right)\right)^{2}}$$. It is calculated using the test data, implying that for each method we obtain 100 RMSE values. Using these RMSE results, we can generate Figure \ref{FIG4-1}.

\begin{table}[!htbp]
	\begin{center}
		\begin{tiny}
			\caption{Comparison of Different Methods under high noise setting}	
			\centering
			\begin{tabular}{|c|c|c|c|c|c|c|c|c|c|} 
				\toprule
				& BART & XBART & XS40T1 & XS40T2 & XS80T1 & XS80T2 & S500 & S1000 & S2000  \\ 
				\hline
				Average Time & 26   & 0.5   & 1.5    & 1.7    & 2.7    & 3.1    & 81   & 151   & 415    \\ 
				\hline
				Ratio        & 17.3 & 0.3   & 1.0    & 1.1    & 1.8    & 2.1    & 54.0 & 100.7 & 276.7  \\
				\bottomrule
			\end{tabular}
		\end{tiny}
	\end{center}
\end{table}

From Figure \ref{FIG4-1} we can see
\begin{itemize}
	\item Under high noise conditions, BART exhibits the poorest estimation accuracy, while XBART shows a noticeable improvement compared to BART.
	\item By initially obtaining the tree structures through the XBART method and then pursuing the optimal window width, the accelerated Bayesian additive regression trees (BART) generated yield similar precision to SBART. With increasing iterations, precision improves, though not dramatically. Likewise, employing a linear gate function may slightly compromise estimation accuracy.
	\item Increasing the iteration count for SBART does not lead to a significant enhancement in estimation accuracy. However, the iterative nature of SBART prevents us from determining an optimal iteration count in advance, necessitating a higher number of iterations to ensure model convergence.
	\item In terms of computational efficiency, considering 40 iterations with accelerated smoothed Bayesian additive regression trees employing a linear gate function as the baseline, BART's time consumption is 17.3 times that of the baseline. SBART's time consumption with the configuration of discarding the first 2000 iterations and retaining the subsequent 2000 iterations is 276.7 times that of the baseline, even if considering 500 iterations, it is 54 times the baseline. It can be asserted that computational efficiency experiences a significant enhancement under this configuration.
	
\end{itemize}

However, it's worth noting that in this scenario of high noise conditions, it's challenging for non-parametric models to capture some local fluctuations. 
Consequently, the simulated tree structures tend to be shallow, and the ASBART only require about 3 times increase in time to complete the smoothing step.

\begin{figure}[!htbp]
	\caption{Comparation of different method in low noise scenario}
	\label{FIG4-2}
	\begin{center}
		\includegraphics[width=0.8\textwidth]{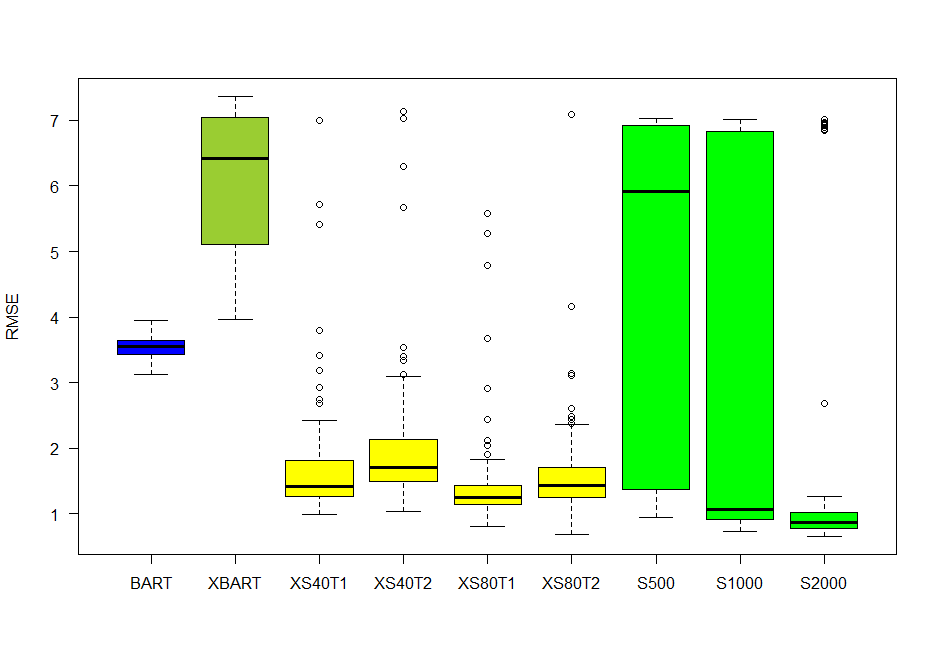}
	\end{center}
\end{figure}

We then focus on the low noise setting.
Figure \ref{FIG4-2} illustrates the performance of different methods under low noise conditions.

\begin{table}[!htbp]
	\begin{center}
		\begin{tiny}
			\caption{Efficiency comparison of different methods under low noise conditions.}	
			\centering
			\begin{tabular}{|c|c|c|c|c|c|c|c|c|c|} 
				\toprule
				& BART & XBART & XS40T1 & XS40T2 & XS80T1 & XS80T2 & S500 & S1000 & S2000  \\ 
				\hline
				Average Time & 26   & 2   & 40    & 48    & 105    & 137    & 90   & 223   & 528    \\ 
				\hline
				Ratio        & 0.6 & 0.1   & 1.0    & 1.2    & 2.6    & 3.4    & 2.2 & 5.5 & 13.1  \\
				\bottomrule
			\end{tabular}
		\end{tiny}
	\end{center}
\end{table}	

\begin{itemize}
	\item Under low noise conditions, BART surprisingly outperforms XBART. This observation is also mentioned in the original XBART paper, where it's noted that BART might perform better than XBART under low noise scenarios.
	\item Notably, the performance of SBART with the retained 500 iterations is comparatively poor. However, with an increase in the number of iterations, there is a substantial improvement in estimation accuracy.
	\item ASBART maintain a clear advantage over both XBART and BART. In comparison to SBART, ASBART yield results that are more stable than S1000, though the superiority of S2000 is more pronounced.
	
	\item Within accelerated smoothed Bayesian additive regression trees, the results of using a linear gate function are better than those of the sigmoid gate function at this stage. This result is reserved for further analysis, and it improves with increased iteration counts.
	\item From a computational efficiency perspective, considering 40 iterations with accelerated smoothed Bayesian additive regression trees employing a linear gate function as the baseline, S1000 requires 5.5 times the time, and S2000 requires 13.1 times the time. Looking from this perspective, the goal of enhancing computational efficiency for smoothed models has been largely achieved. However, there is a slight compromise in estimation accuracy compared to S2000. In contrast to the low noise scenario, the time consumption increases significantly at this point due to the deeper tree structures, which in turn increase the time required for optimal window width search.
	
\end{itemize}

\section{Conclusion and Looking Forward}
\label{sec:5}

By integrating the strengths of the XBART model into the SBART framework, we have significantly expedited computational speed while maintaining a comparable level of precision to SBART. The feasibility of this algorithm is demonstrated through empirical experiments.

In contrast to SBART, the main distinction here is the adoption of a grid-point search method for optimal window width, which can also be substituted with randomized sampling of window widths. Additionally, unlike SBART, our model does not preprocess covariates into percentile data. Instead, it calculates comparable window widths for each covariate post-process, which may result in the drawback of potential incomparability among covariates.

Furthermore, during iterations, we employ the Metropolis-Hastings (MH) algorithm to select trees. However, this may lead to results becoming trapped in local optima, causing incremental changes in tree structures as iterations progress, requiring a substantial number of iterations to escape such a scenario. To address this issue, a potential solution is to decrease the number of attempted window configurations with each iteration and introduce a restarting process between iterations, as later mentioned in subsequent chapters. This could effectively force the algorithm to break free from local optima.

As seen earlier, in a distributed environment, applying SBART involves a substantial number of ineffective attempts, accompanied by extensive information exchange. Leveraging XBART's rapid convergence characteristics in such an environment can significantly enhance computational efficiency on the basis of distribution. This entails performing candidate model computations and window width selection locally, followed by the selection of different smoothed decision trees on various worker nodes. Distributing the selected tree structures to individual nodes through this process could further accelerate distributed SBART computation. Additionally, this approach markedly reduces the amount of necessary information exchange.

\vskip20pt
\def\refhg{\hangindent=20pt\hangafter=1}
\def\refmark{\par\vskip 1mm\noindent\refhg}
\bibliographystyle{plainnat}

\bibliography{RF}

\end{document}